\documentclass{IEEEtran}
\usepackage{spconf,amsmath,graphicx}
\usepackage{tabularx}  
\usepackage{booktabs}  
\usepackage{float}     
\usepackage{stfloats}  
\usepackage{caption}
\usepackage[hyphens]{url}
\title{An Agentic AI Workflow for Liver Lesion Segmentation and Structured Radiology Reporting}

\name{
\begin{tabular}{ccc}
\shortstack{
Fatma Ahmed Mansour\\
\small Queen's University, Canada\\
\small 25nyln@queensu.ca
}
&
\shortstack{
Rahatara Ferdousi\\
\small Queen's University, Canada\\
\small rahatara.ferdousi@queensu.ca
}
&
\shortstack{
Zyad Medhat Mahmoud Ashor\\
\small Faculty of Engineering, Mansoura Univ.\\
\small zyadashor766@gmail.com
}
\\[2.5ex]
\shortstack{
Manar Fathy Elshenawy\\
\small Queen's University, Canada\\
\small 25ssyn@queensu.ca
}
&
\shortstack{
Ibrahim Shehaby\\
\small RadioDiagnosis Consultant, Rology Inc.\\
\small ibrahim.shehaby@rology.net
}
&
\shortstack{
Salma Elmongi\\
\small AI Engineer, Rology Inc.\\
\small salma.elmongi@rology.net
}
\\[2.5ex]
\multicolumn{3}{c}{
\shortstack{
S. Ahmed\\
\small Math. Dept., Faculty of Science, Cairo University, Egypt\\
\small sameh@sci.cu.edu.eg
}
}
\end{tabular}
}
\address{}

\begin{document}
%
\maketitle

\begin{abstract}
Despite advances in automated liver lesion segmentation, a gap remains between image-derived predictions and clinically interpretable radiology reports. Bridging this gap requires quantitative measurements that are both reproducible and traceable.
We present an evidence-grounded agentic AI framework
that connects liver and lesion segmentation with
deterministic quantification and structured reporting.
TotalSegmentator-based nnU-Net models delineate the
liver, Couinaud segments, and lesions, while computational
tools extract volumes, dimensions, counts, and anatomical
locations. A large language model (LLM) organizes these
measurements into a predefined reporting schema, with
numerical consistency checks against source measurements.
On 51 LiTS cases, the segmentation stage achieved a
liver Dice score of 0.947, lesion sensitivity of 75.6\%,
and positive predictive value of 68.9\%.
Exploratory screening of 25,494 MERLIN reports identified
10,208 (40.0\%) with liver-related abnormalities. By connecting quantitative imaging evidence with agentic report generation, this work takes a step toward human-centered medical AI that prioritizes traceability and clinical relevance.

\end{abstract}


\vspace{-0.3cm}

\section{Introduction}
\label{sec:intro}

Quantitative assessment of liver lesions in computed
tomography (CT) is essential for treatment planning
and disease monitoring. Radiologists evaluate lesion
size, volume, and anatomical location to assess
disease progression and treatment response. However,
manual delineation and measurement remain
time-consuming and subject to observer variability.

Recent advances in deep learning have enabled
automated liver and lesion segmentation using
frameworks such as nnU-Net \cite{isensee2021nnu} and
TotalSegmentator \cite{wasserthal2023totalsegmentator}. Related studies have
explored Couinaud segment delineation \cite{mathai2026couinaud} and
semi-automated lesion quantification with
structured reporting \cite{jobstl2025lirads}. Despite these advances,
integrating segmentation-derived measurements
into structured reports while preserving their
traceability remains an important challenge.
This is particularly relevant to large language
models (LLMs), which may generate numerical
findings unsupported by image-derived evidence.

To address this challenge, we propose an
evidence-grounded agentic AI framework that
integrates liver lesion segmentation,
deterministic quantitative analysis, and
structured radiology reporting. The framework
derives lesion measurements and anatomical
locations from volumetric masks, while an
LLM agent organizes these findings into a
predefined reporting schema. Numerical
consistency checks link reported values
to their computational sources, supporting
transparent and radiologist-reviewable reporting.

The main contributions are:
\begin{itemize}
    \item An integrated agentic workflow
    connecting automated liver segmentation,
    quantitative analysis, and structured
    radiology reporting.

    \item A deterministic measurement pipeline
    with numerical consistency checks to
    support traceable reporting of lesion
    characteristics.

    \item An experimental assessment of
    segmentation and lesion detection on LiTS,
    complemented by exploratory screening
    of liver-related findings in MERLIN reports.
\end{itemize}

The source code is publicly available at
\url{https://github.com/Xfatmaah/AI-Assisted-Liver-Lesion-Segmentation-for-Structured-Radiology-Reporting},
and a demonstration video is available at
\url{https://www.youtube.com/watch?v=4Wm3c_QAp38}.

\section{Related Work}
\label{sec:related}

\subsection{Liver Segmentation and Quantitative Analysis}

The self-configuring nnU-Net framework \cite{isensee2021nnu} and
TotalSegmentator \cite{wasserthal2023totalsegmentator,dantonoli2025totalsegmentator} have advanced automated
anatomical segmentation in CT and MRI.
Nicoli et al. \cite{nicoli2025liver} extended these approaches to
liver segments and lesions, including quantitative
assessment of hepatocellular carcinoma.
The LiTS benchmark \cite{bilic2023lits} provides a reference
for liver and tumor segmentation evaluation.
Automated Couinaud delineation \cite{mathai2026couinaud} and
semi-automated HCC quantification \cite{jobstl2025lirads} further
demonstrate the role of segmentation in
quantitative liver assessment.

\subsection{LLM-Assisted Radiology Reporting}

Recent studies have explored LLM-based structured
radiology reporting \cite{busch2025llm,delbrouck2025srrg,csrrg2025}, evidence-driven
report generation \cite{eviagent2025}, uncertainty-aware
labeling \cite{uncertainty2025agent}, and interpretable report
evaluation \cite{icare2025eval}. These approaches address
report generation, evidence integration,
and assessment from different perspectives.

Building on these advances, our work integrates
segmentation-derived quantitative measurements
with constrained LLM-based report composition.
Unlike direct numerical generation by an LLM,
measurements are computed using deterministic
tools and checked against their source outputs.
The contribution lies in this integrated,
traceable workflow rather than a new
segmentation model or reporting algorithm.

\section{Proposed Approach}
\label{sec:methodology}

\subsection{System Architecture Overview}

\begin{figure*}
    \centering
    \includegraphics[width=0.85\textwidth]{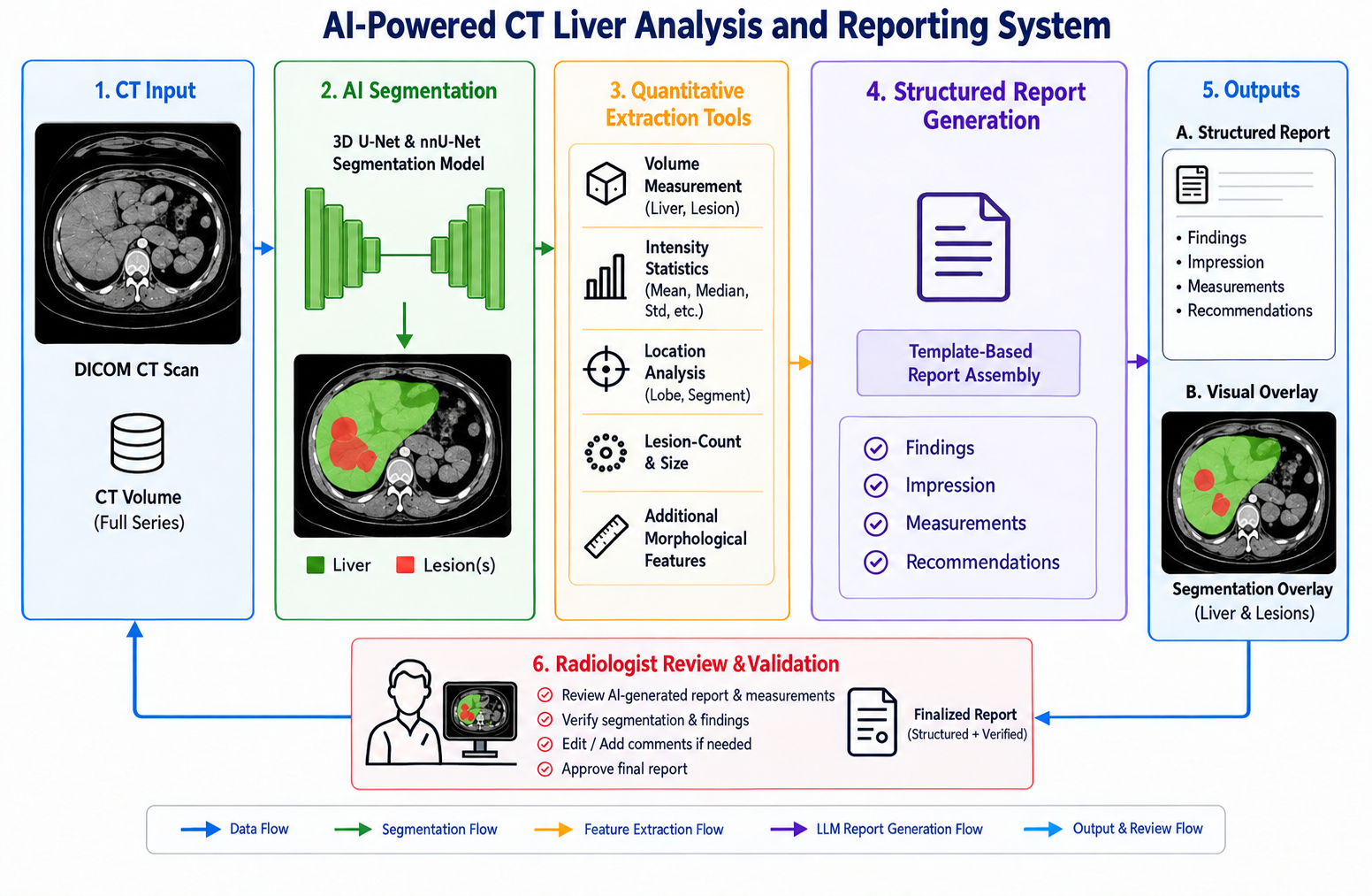}
    \caption{Overview of the proposed agentic AI workflow}
    \label{fig:workflow}
\end{figure*}

The proposed framework follows a five-stage agentic
workflow, as illustrated in Fig.~\ref{fig:workflow}.

To reduce unsupported numerical generation, the
LLM-based report composer operates on outputs from
deterministic measurement tools rather than estimating
quantities directly from CT images. Each reported
measurement is linked to its computational source,
enabling verification and radiologist review.

\subsection{Segmentation Stage}

Contrast-enhanced CT scans are processed using
pre-trained nnU-Net models from the TotalSegmentator
framework. The models generate segmentation masks
for the liver parenchyma, individual Couinaud
anatomical segments, and focal hepatic lesions.

\begin{figure}[H]
    \centering
    \includegraphics[width=\linewidth]{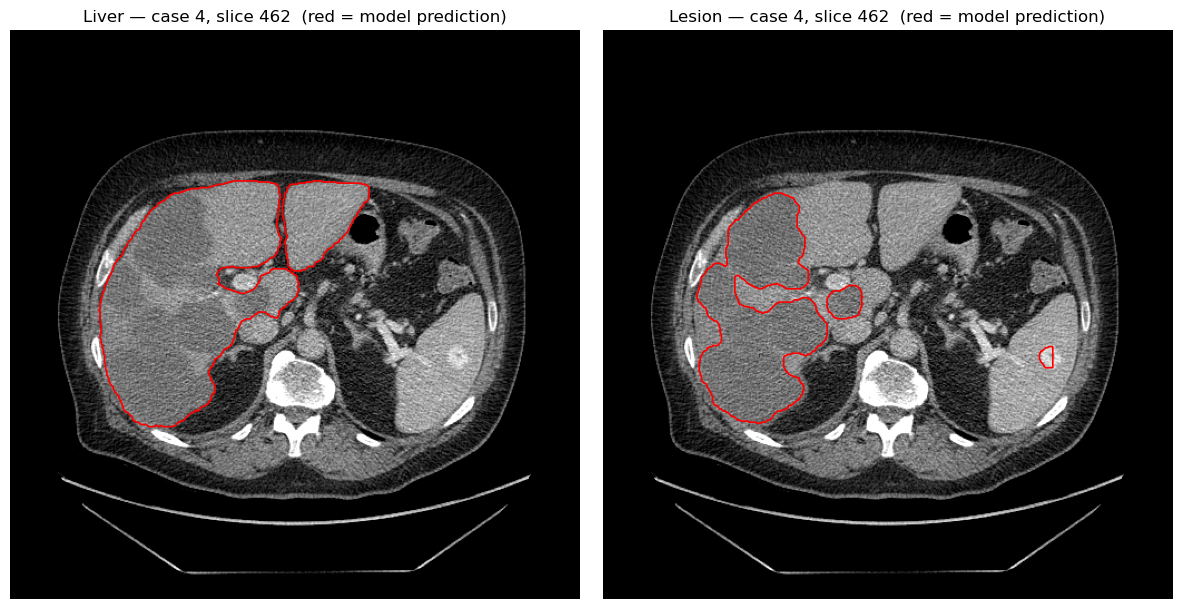}
    \caption{Representative CT segmentation masks
    illustrating the delineated liver and lesions.}
    \label{fig:segmentation}
\end{figure}

\subsection{Deterministic Extraction Tools}

A Python-based computational layer extracts
quantitative measurements from the predicted
three-dimensional segmentation masks:

\begin{enumerate}
    \item \textbf{Liver and segment volumes:}
    Computed from voxel counts and physical voxel
    spacing as $V=N_v s_x s_y s_z/1000$ mL,
    where spacing is expressed in mm.

    \item \textbf{Lesion volumes:}
    Total and individual lesion volumes are
    computed using the same volumetric formula.

    \item \textbf{Lesion count:}
    Determined through 3D connected-component
    labeling of lesion masks.

    \item \textbf{Lesion dimensions:}
    Estimated from the X-, Y-, and Z-axis
    extents of each lesion's bounding box (mm),
    rather than true Feret diameters.

    \item \textbf{Couinaud localization:}
    Each lesion is assigned to the anatomical
    segment with the largest mask overlap.
\end{enumerate}

The extraction tools are exposed to the agent
through a lightweight FastAPI backend.

\subsection{Diagnostic Findings from RADAR}

For each CT scan, RADAR estimates probabilities
for 18 selected liver-related findings from
its 146 supported categories \cite{radar2026}. These
predictions provide supplementary diagnostic
context alongside segmentation-derived
measurements but do not modify the computed
quantitative values.

\subsection{Evidence-Grounded Report Generation
and Verification}

The LLM agent receives structured measurement
data in JSON format and accesses predefined
API tools. Report generation follows a fixed
schema containing study metadata, anatomical
locations, volumetric measurements, linear
dimensions, and fields for longitudinal changes.

Two verification steps are applied:
(i) schema validation to check output structure
and (ii) numerical consistency checking against
the source measurement JSON. Values without
a corresponding computational source are flagged
for review. These checks establish numerical
traceability but do not validate segmentation
accuracy. Longitudinal fields are retained
in the schema but are not evaluated because
serial CT examinations are unavailable.

\subsection{Radiologist Review Interface}

The structured report is displayed alongside
segmentation overlays aligned with axial CT
slices. The interface allows radiologists to
inspect image-derived findings, review
measurements, and modify the report before
final approval.

\section{Experimental Setup}

\begin{itemize}
    \item \textbf{Datasets:}
    The LiTS dataset \cite{bilic2023lits} was used to evaluate
    liver and lesion segmentation on 51 CT cases.
    MERLIN radiology reports were used for
    exploratory screening of liver abnormalities.

    \item \textbf{Evaluation Metrics:}
    Segmentation was assessed using the Dice
    similarity coefficient (DSC). Lesion detection
    was evaluated using sensitivity, positive
    predictive value (PPV), and false positives
    per case (FP/c).

    \item \textbf{Baseline Comparison:}
    Results were compared with published findings
    from Nicoli et al~\cite{nicoli2025liver}. Differences in
    evaluation cohorts and protocols limit
    direct comparability.

    \item \textbf{Reporting Evaluation:}
    The reporting prototype incorporates numerical
    consistency checks against computed measurements.
    Reporting accuracy and clinical usability
    were not quantitatively evaluated.
\end{itemize}

\section{Results}
\label{sec:results}

\subsection{Segmentation Performance}

\begin{figure}[H]
    \centering
    \includegraphics[width=\linewidth]{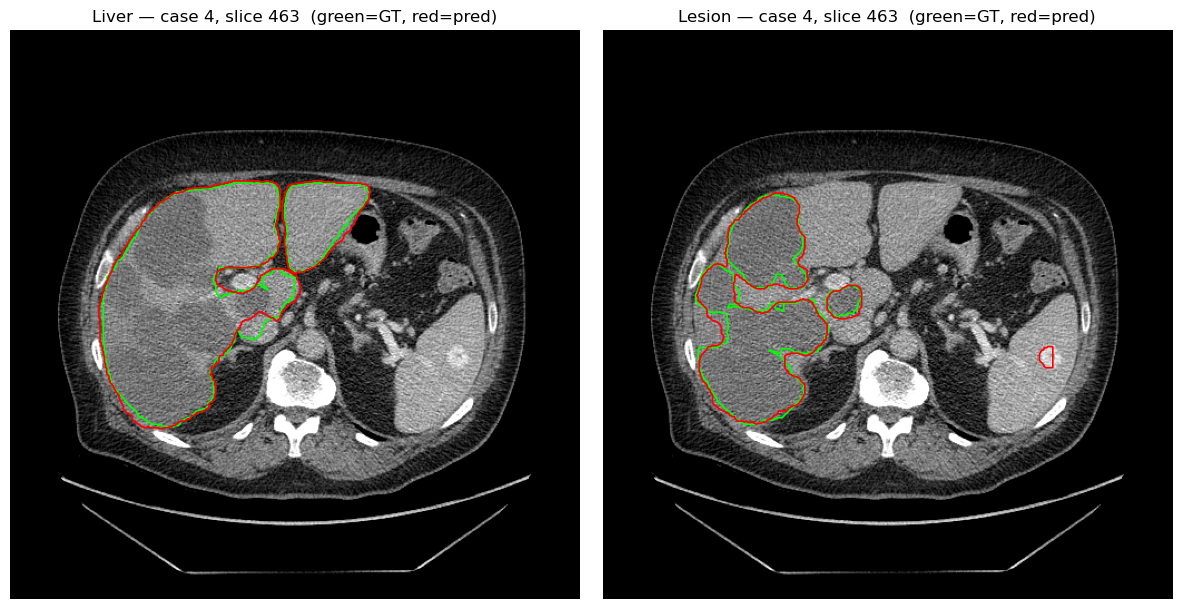}
    \caption{Representative LiTS CT segmentation
    overlay showing ground-truth-only regions
    (green) and prediction-only regions (red).}
    \label{fig:seg_overlay}
\end{figure}

All 51 evaluated LiTS cases passed the orientation
sanity check and were successfully processed,
with no flagged or failed cases. Fig.~\ref{fig:seg_overlay}
illustrates a representative comparison between
predicted and reference liver and lesion masks.

The overlay highlights regions of agreement and
disagreement, providing a qualitative assessment
of segmentation boundaries. The segmentation stage achieved a liver Dice
coefficient of 0.947, indicating substantial
spatial overlap with reference annotations.
Lesion segmentation was more challenging,
with a reported Dice score of 0.63.

\subsection{Lesion Detection Performance}

Table~\ref{tab:lits_comparison} summarizes
lesion-level performance alongside the
published results of Nicoli et al. ~\cite{nicoli2025liver}. The evaluated model achieved a lesion
sensitivity of 75.6\%, PPV of 68.9\%,
and 0.70 false positives per case.
Compared with the published reference,
sensitivity was 9.9 percentage points
higher (75.6\% versus 65.7\%), while
PPV was 11.27 percentage points lower
(68.9\% versus 80.17\%). False positives
per case were similar (0.70 versus 0.73).

These results indicate a trade-off between
lesion detection sensitivity and precision.
Although more reference lesions were detected
proportionally, a smaller proportion of
predicted lesions corresponded to reference
findings. Because the evaluation cohorts
and protocols were not identical, these
differences should not be interpreted as
evidence of superior model performance.

\begin{table}[t]
\centering
\caption{Lesion detection and segmentation
results on LiTS. Published reference values
are from Nicoli et al.~\cite{nicoli2025liver}; results are
not from a paired evaluation.}
\label{tab:lits_comparison}
\resizebox{\columnwidth}{!}{%
\begin{tabular}{lcc}
\toprule
Metric & Nicoli et al.~\cite{nicoli2025liver} & Ours \\
\midrule
Lesion sensitivity (\%) & 65.7 & 75.6 \\
PPV (\%)                & 80.17 & 68.9 \\
FP per case             & 0.73 & 0.70 \\
Lesion Dice             & 0.59 & 0.63 \\
\bottomrule
\end{tabular}%
}
\end{table}

\begin{figure}[t]
    \centering
    \includegraphics[width=\linewidth]
    {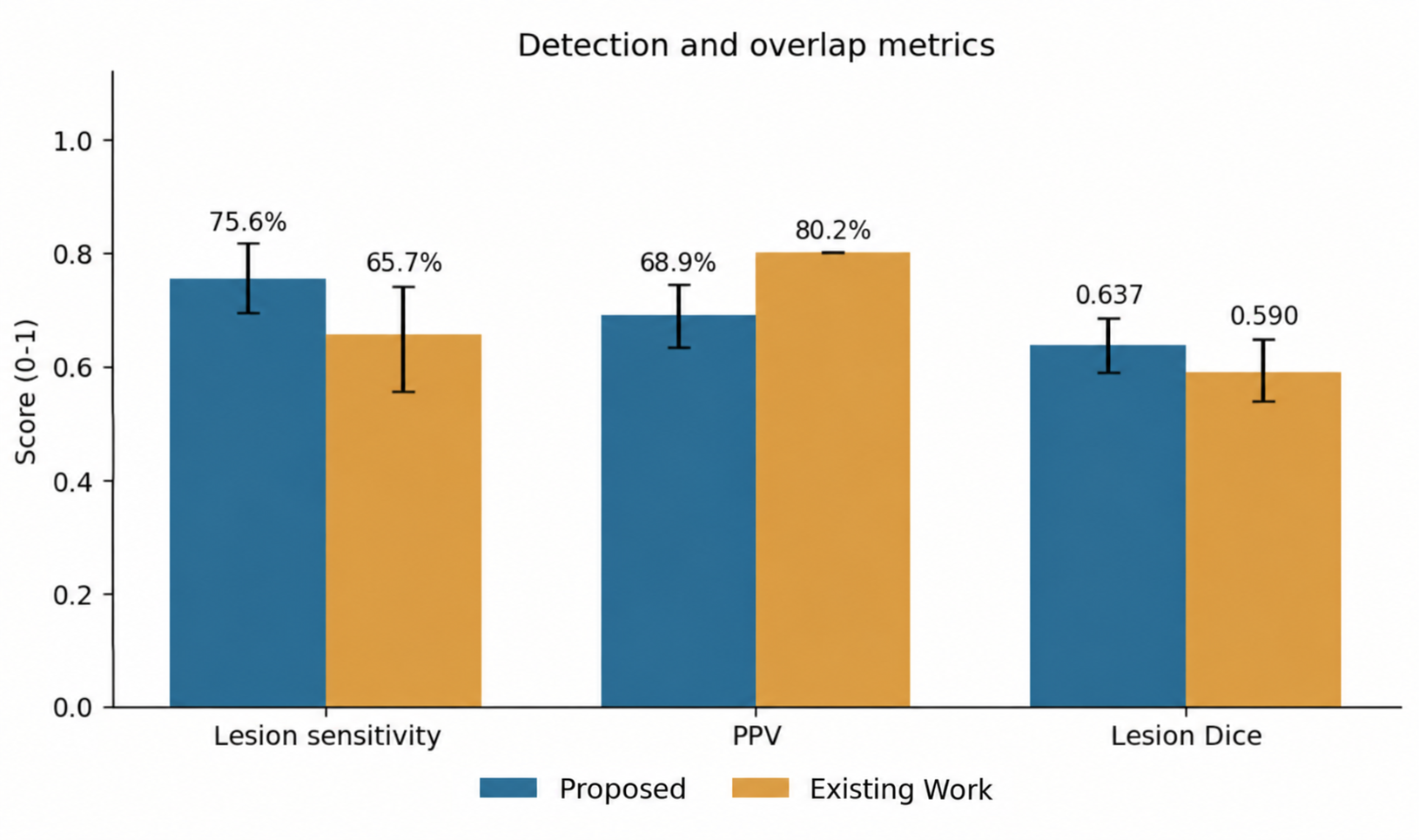}
    \caption{Comparison of lesion detection
    and segmentation metrics with the
    published reference results~\cite{nicoli2025liver}.
    Differences are descriptive and do
    not represent a controlled comparison.}
    \label{fig:evaluation}
\end{figure}

\subsection{Exploratory Report Screening}

The report-screening stage processed
25,494 MERLIN radiology reports and
identified 10,208 (40.0\%) as containing
liver-related abnormalities. These findings
describe the distribution of predicted
screening outcomes. Classification accuracy
cannot be established without comparison
against independently verified labels.

\section{Discussion}
\label{sec:discussion}
The LiTS results indicate reliable liver segmentation,
while lesion detection remains more challenging, as
reflected in the sensitivity--PPV trade-off. These
findings support the use of automated segmentation
as the imaging foundation of the proposed framework.
The primary contribution lies in connecting these
outputs to deterministic lesion quantification and
evidence-grounded structured reporting, with numerical
verification designed to preserve measurement
traceability.

The current evaluation is limited to 51 LiTS cases,
with published results used for reference rather
than direct comparison. Future work will extend
evaluation to larger, matched cohorts, assess
measurement accuracy against reference annotations,
and quantify reporting consistency and radiologist
usability. These investigations will establish
the extent to which the integrated workflow can
support reproducible quantitative radiology reporting.

\section{Conclusion}
\label{conclusion}
This work presents an evidence-grounded agentic AI
framework that connects automated liver and lesion
segmentation with quantitative analysis and structured
radiology reporting. The LiTS evaluation demonstrates
the performance of the underlying segmentation stage,
while the proposed architecture separates image-derived
measurements from LLM-based report composition.
By integrating deterministic measurement tools,
numerical consistency checks, and radiologist review,
the framework provides a traceable approach to
organizing quantitative imaging findings.
Future work will validate measurement accuracy,
reporting consistency, and clinical usability
to assess its potential for reliable,
human-supervised radiology workflows.

\section*{Ethics Statement}

This study uses publicly available, de-identified
LiTS and MERLIN datasets and involves no new
patient data collection or re-identification.
No additional institutional ethical approval
was required. The proposed system is a research
prototype intended for radiologist-supervised
decision support, not autonomous diagnosis or
clinical deployment.

\end{document}